\documentclass[11pt]{article}
\usepackage{hyperref}
\usepackage{url}
\usepackage{amsmath}
\usepackage[pdftex]{graphicx}
\usepackage{cite}
\usepackage{color}
\usepackage[margin=1in]{geometry}

\begin{document}

\title{A Portable, Self-Contained Neuroprosthetic Hand with Deep Learning-Based Finger Control}

\author{Anh Tuan Nguyen $^{*,1,5}$, Markus W. Drealan$^{*,1}$, Diu Khue Luu$^{1}$, Ming Jiang$^{2}$, \\Jian Xu$^{1}$, Jonathan Cheng$^{3}$, Qi Zhao$^{2}$, Edward W. Keefer$^{4,5}$, and Zhi Yang$^{1,5}$}
\date{	\raggedright $^{*}$ Co-first authors \\
			\raggedright $^{1}$ Biomedical Engineering, University of Minnesota, Minneapolis, MN, USA \\
			\raggedright $^{2}$ Computer Science and Engineering, University of Minnesota, Minneapolis, MN, USA \\
			\raggedright $^{3}$ Plastic Surgery, University of Texas Southwestern Medical Center, Dallas, TX, USA \\
			\raggedright $^{4}$ Nerves Incorporated, Dallas, TX, USA \\
			\raggedright $^{5}$ Fasikl Incorporated, Minneapolis, MN, USA \\
			\raggedright Correspondence: A. T. Nguyen (Email: nguy2833@umn.edu) and M. W. Drealan (Email: dreal003@umn.edu)\\}
\maketitle

\begin{abstract}
\textit{Objective}: Deep learning-based neural decoders have emerged as the prominent approach to enable dexterous and intuitive control of neuroprosthetic hands. Yet few studies have materialized the use of deep learning in clinical settings due to its high computational requirements.
\textit{Methods}: Recent advancements of edge computing devices bring the potential to alleviate this problem. Here we present the implementation of a neuroprosthetic hand with embedded deep learning-based control. The neural decoder is designed based on the recurrent neural network (RNN) architecture and deployed on the NVIDIA Jetson Nano - a compacted yet powerful edge computing platform for deep learning inference. This enables the implementation of the neuroprosthetic hand as a portable and self-contained unit with real-time control of individual finger movements.
\textit{Results}: The proposed system is evaluated on a transradial amputee using peripheral nerve signals (ENG) with implanted intrafascicular microelectrodes. The experiment results demonstrate the system's capabilities of providing robust, high-accuracy (95-99\%) and low-latency (50-120 msec) control of individual finger movements in various laboratory and real-world environments.
\textit{Conclusion}: Modern edge computing platforms enable the effective use of deep learning-based neural decoders for neuroprosthesis control as an autonomous system.
\textit{Significance}: This work helps pioneer the deployment of deep neural networks in clinical applications underlying a new class of wearable biomedical devices with embedded artificial intelligence. 

%Objective: Deep learning-based neural decoders have emerged as the prominent approach to enable dexterous and intuitive control of neuroprosthetic hands. Yet few studies have materialized the use of deep learning in clinical settings due to its high computational requirements. Methods: Recent advancements of edge computing devices bring the potential to alleviate this problem. Here we present the implementation of a neuroprosthetic hand with embedded deep learning-based control. The neural decoder is designed based on the recurrent neural network (RNN) architecture and deployed on the NVIDIA Jetson Nano - a compacted yet powerful edge computing platform for deep learning inference. This enables the implementation of the neuroprosthetic hand as a portable and self-contained unit with real-time control of individual finger movements. Results: The proposed system is evaluated on a transradial amputee using peripheral nerve signals (ENG) with implanted intrafascicular microelectrodes. The experiment results demonstrate the system's capabilities of providing robust, high-accuracy (95-99%) and low-latency (50-120 msec) control of individual finger movements in various laboratory and real-world environments. Conclusion: Modern edge computing platforms enable the effective use of deep learning-based neural decoders for neuroprosthesis control as an autonomous system. Significance: This work helps pioneer the deployment of deep neural networks in clinical applications underlying a new class of wearable biomedical devices with embedded artificial intelligence. 
\end{abstract}


\section{Introduction}
\begin{figure}[p]
\centering
\includegraphics[trim=5 0 0 0, clip=true, width=0.8\textwidth]{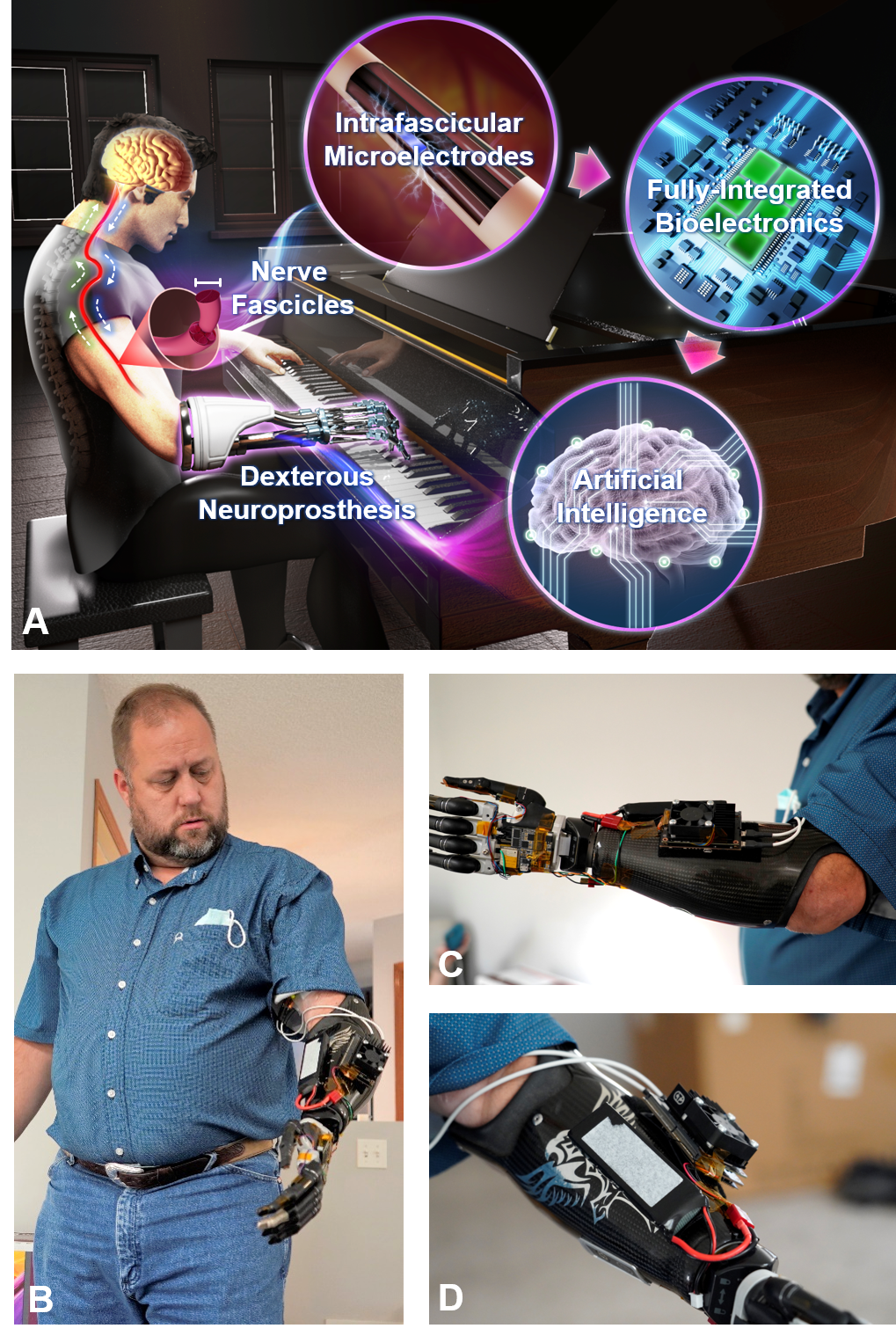}
%\vspace{-5pt}
\caption{(A) Artistic concept of the next-generation dexterous and intuitive neuroprostheses (Figure adopted from \cite{2020_Nguyen}). (B, C, D) Here we present the implementation of a functional, fully portable, self-contained prototype toward this goal.}
%\vspace{-15pt}
\label{Fig_Concept}
\end{figure}

Applications of deep learning to analyze, interpret, and decode biomedical data has been gaining steady momentum in recent years \cite{2018_Mahmud, 2019_Esteva}. In the rapidly developing neuroprosthetics field, deep learning-based neural decoders have emerged as the prominent approach to facilitate the next-generation dexterous and intuitive neuroprostheses (Fig. \ref{Fig_Concept}) \cite{2012_Sussillo, 2016_Atzori, 2018_George, 2019_Alazrai, 2019_Dantas, 2020_Nguyen, 2021_Luu}. Deep learning-based artificial intelligence (AI) could be the missing link that would allow users to harness the full range of movements of dexterous prosthetic systems like the DEKA Arm \cite{2014a_Resnik, 2014b_Resnik}, the APL ARM \cite{2011_Johannes}, and the DLR Hand Arm system \cite{2011_Grebenstein, 2012_Grebenstein}.

Several previous studies have demonstrated the superior efficacy of deep learning approaches compared to conventional algorithms for decoding human motor intent from neural data. Atzori \textit{et al.} (2016) \cite{2016_Atzori} explore the use of convolutional neural networks (CNN) to classify more than 50 hand movements using surface electromyography (sEMG) data obtained from intact and amputee subjects. George \textit{et al.} (2018) \cite{2018_George} apply CNN to enable an amputee to control a multi-degrees of freedom (DOF) prosthetic hand in real-time using peripheral nerve and intramuscular electromyography (iEMG) data. Our recent work Nguyen \& Xu \textit{et al.} (2020) \cite{2020_Nguyen} compares the performance of deep learning decoders based on CNN and recurrent neural network (RNN) against classic machine learning techniques, including multilayer perceptron (MLP), random forest (RF), and support vector machine (SVM). We show that deep learning approaches outperform other techniques across all performance metrics in decoding dexterous finger movements using peripheral nerve data acquired from an amputee. The results are further cemented in Luu \& Nguyen \textit{et al.} (2021) \cite{2021_Luu} where we investigate different strategies to optimize the efficacy of the deep learning motor decoder.

While these studies show promising results, the application of deep learning on portable devices for long-term clinical uses remains challenging \cite{2017_Ghafoor, 2019_Wolf}. The efficacy of deep learning comes at the cost of computational complexity. It is well-known that running deep learning models on conventional central processing units (CPU) found on most low-power platforms is hugely inefficient. In practice, the vast majority of deep learning models must be trained and deployed using graphics processing units (GPU) which have hundreds to thousands of multi-threaded computing units specialized for parallelized floating-point matrix multiplication. As a result, until neuromorphic chips like the IBM TrueNorth \cite{2016_Nurse} become available, many portable neuroprosthetic systems have to divert to less computational techniques like K-nearest neighbor (KNN) \cite{2011a_Cipriani, 2019_Risso}, Kalman filter \cite{2020_George, 2020_Brinton} or variants of artificial neural networks (ANN) \cite{2019_Furui}.
     
Recent advancements of edge computing devices bring the potential to alleviate this problem. Early edge computing devices focused on general-purpose applications, but recently, there has been an increase in the development of compact hardware for deep learning uses. One class of such hardware includes inference-only, USB-powered devices like the Intel Neural Compute Stick 2\footnote{\url{https://software.intel.com/content/www/us/en/develop/hardware/neural-compute-stick.html}}, and the Google Coral Accelerator\footnote{\url{https://coral.ai/products/accelerator}}. Both use proprietary integrated circuits that perform deep learning inference with ultra-low power consumption ($<$1 W). Unfortunately, the current software support is limited to highly customized neural networks, which hinders the full potential of our neural decoder implementation based on RNN.

Another class of edge computing devices for deep learning comprises of standalone computers with integrated GPU. The most well-known example is the NVIDIA Jetson family\footnote{\url{https://www.nvidia.com/en-us/autonomous-machines/embedded-systems}} which includes power-efficient, general-purpose mini-computers with CPU, GPU, RAM, flash storage, etc. specifically designed to deploy AI in autonomous applications. While these devices consume considerably more power (5-20 W), the current software offers full support for Compute Unified Device Architecture (CUDA) cores, allowing direct uses of the most popular deep learning libraries like TensorFlow, PyTorch, and Caffe. This offers the most appropriate trade-off among size, power, and performance for our neural decoder implementation.

This paper addresses the challenge of efficiently deploying deep learning neural decoders on a portable, edge computing platform, translating previous benchtop motor decoding experiments into real-life applications toward long-term clinical uses. We chose the Jetson Nano module as the main processing unit that handles all data acquisition, data processing, and deep learning-based motor decoding. We also design customized printed circuit boards (PCB) to support the Jetson Nano's function and operate the robotic hand. The entire system is small enough to be comfortably attached to the existing prosthetic limb of a transradial amputee.

To the best of our knowledge, this system is one of the first to efficiently implement deep learning neural decoders in a portable platform for clinical neuroprosthetic applications. It is feasible thanks to multiple innovations that we have pioneered across various system's components. The first innovation lies in the development of the intrafascicular microelectrode array that connects nerve fibers and bioelectronics, as presented in Overstreet \textit{et al.} (2019) \cite{2019_Overstreet}. The second innovation lies in the design of the Neuronix neural interface microchips that allows simultaneous neural recording and stimulation, as presented in Nguyen \& Xu \textit{et al.} (2020) \cite{2020_Nguyen} and Nguyen \textit{et al.} (2021) \cite{2021_Nguyen}. The third innovation lies in the optimization of the deep learning motor decoding paradigm that helps reduce the decoder's computational complexity, as presented in Luu \& Nguyen \textit{et al.} (2021) \cite{2021_Luu}. The final innovation lies in the implementation of software and hardware based on a state-of-the-art edge computing platform that could support real-time motor decoding, as would be presented in the rest of this manuscript.

%The performance of the neural decoders based on RNN is evaluated on able participants using surface electroneurography/electromyography (ENG/EMG) recordings, and a transradial amputee using peripheral nerve recordings with implanted intrafascicular microelectrodes. Both benchmarks show the deep learning-based decoders could predict the fingers' movement in real-time with high accuracy ranging from 97-99\% for able participants and 95-96\% for the amputee. The total time latency from data acquisition to final prediction output varies from 50msec to 130msec depending on the number of neural decoder models needed. deep learning inference accounts for over 90\% of the module's power consumption and a latency of 30 msec with one model to 100 msec with five models. Furthermore, we show that the neuroprosthesis's functions are robust when the amputee moves among various laboratory and real-world environments, regardless of the arm/body postures.

\section{System Implementation}
\subsection{System Overview}
\begin{figure}[p]
\centering
\includegraphics[width=1\textwidth]{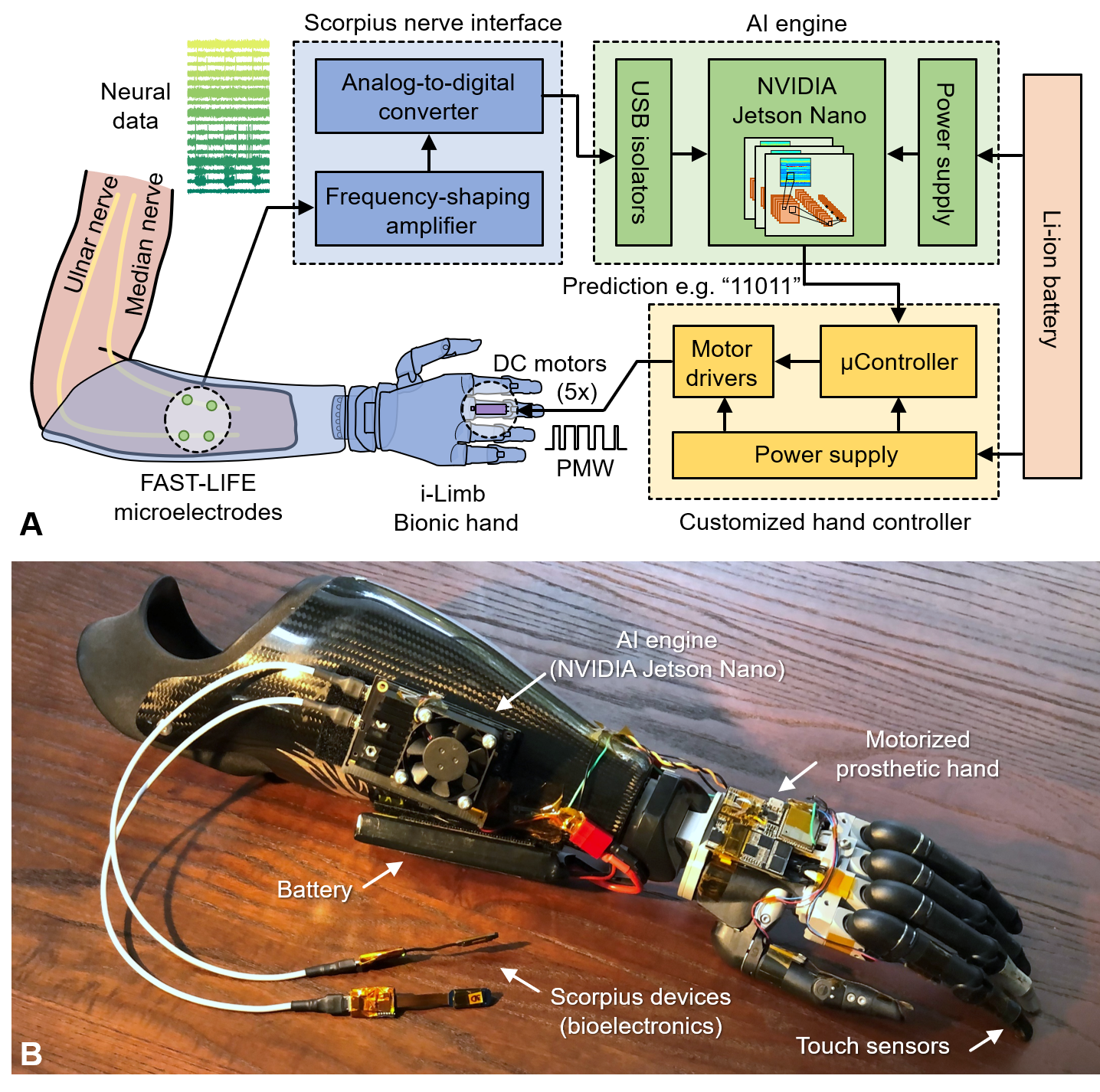}
%\vspace{-5pt}
\caption{(A) Overview of the proposed system, including the prosthetic hand, nerve interface, and deep learning-based neural decoder. (B) A prototype system attached to the amputee's existing prosthetic socket as a self-contained, portable unit.}
%\vspace{-10pt}
\label{Fig_Overview}
\end{figure}

Fig. \ref{Fig_Overview}(A) shows the overview of the proposed neuroprosthetic hand neural decoder. It includes the Scorpius neural interface, the Jetson Nano with a customized carrier board, the customized hand controller, and a rechargeable Li-ion battery. Fig. \ref{Fig_Overview}(B) gives an overall look at the prototype system being attached to the amputee's existing prosthetic socket as a self-contained, portable unit. The system adds additional weights of 90g for the AI engine and 120g for a 7.4V 2,200mAh Li-ion battery. In practice, the entire system could be integrated into the socket's interior, replacing the prosthesis's existing EMG sensors and electronics, thus having minimal impact on the hand's weight and aesthetic.

Nerve data are acquired by our previously proposed Scorpius neural interface \cite{2020_Nguyen}. Each Scorpius device has eight recording channels equipped with the frequency-shaping (FS) amplifier and high-precision analog-to-digital converters (ADC). Several devices can be deployed depending on the number of channels required. The FS neural recorders are proven to be capable of obtaining ultra-low noise nerve signals while suppressing undesirable artifacts \cite{2013a_Xu, 2014_Xu, 2016_Yang, 2018a_Xu, 2018b_Xu, 2018_Yang, 2020_Yang, 2020b_Xu}. Raw nerve data are directly streamed to the Jetson Nano for further processing. 

The core of the system is the AI-engine powered by the Jetson Nano platform. We design a customized carrier board to provide the power management and I/O connectivity for the Nano module. The module is a mini-computer equipped with the Tegra X1 system-on-chip (SoC) that has a Quad-Core ARM Cortex-A57 CPU and a 128-core NVIDIA Maxwell GPU. The GPU has 472 GFLOPs of computational power available for deep learning inference. The module can operate in the 10W mode (4-core CPU 1900 MHz, GPU 1000 GHz) or the 5W mode (2-core CPU 918 MHz, GPU 640 MHz), which translates to about 2h and 4h of continuous usage with the current battery size, respectively. Here fully-trained deep learning models are deployed to translate nerve signals to the subject's true intentions of individual finger movements in real-time. The final predictions are sent over to the hand controller to actuate the prosthetic hand. 

The hand itself is based on the i-Limb platform (TouchBionics/\"{O}ssur, Iceland) with five individually actuated fingers. We replace the i-Limb default driver with a customized hand controller that directly operates the DC motors hidden in each finger according to the deep learning models’ prediction. The controller is designed around the ESP32 module (Espressif Systems, Shanghai, China) with a low-power microcontroller.

\subsection{Hardware Implementation}
\begin{figure}[ht]
\centering
\includegraphics[trim=10 10 0 0, clip, width=1\linewidth]{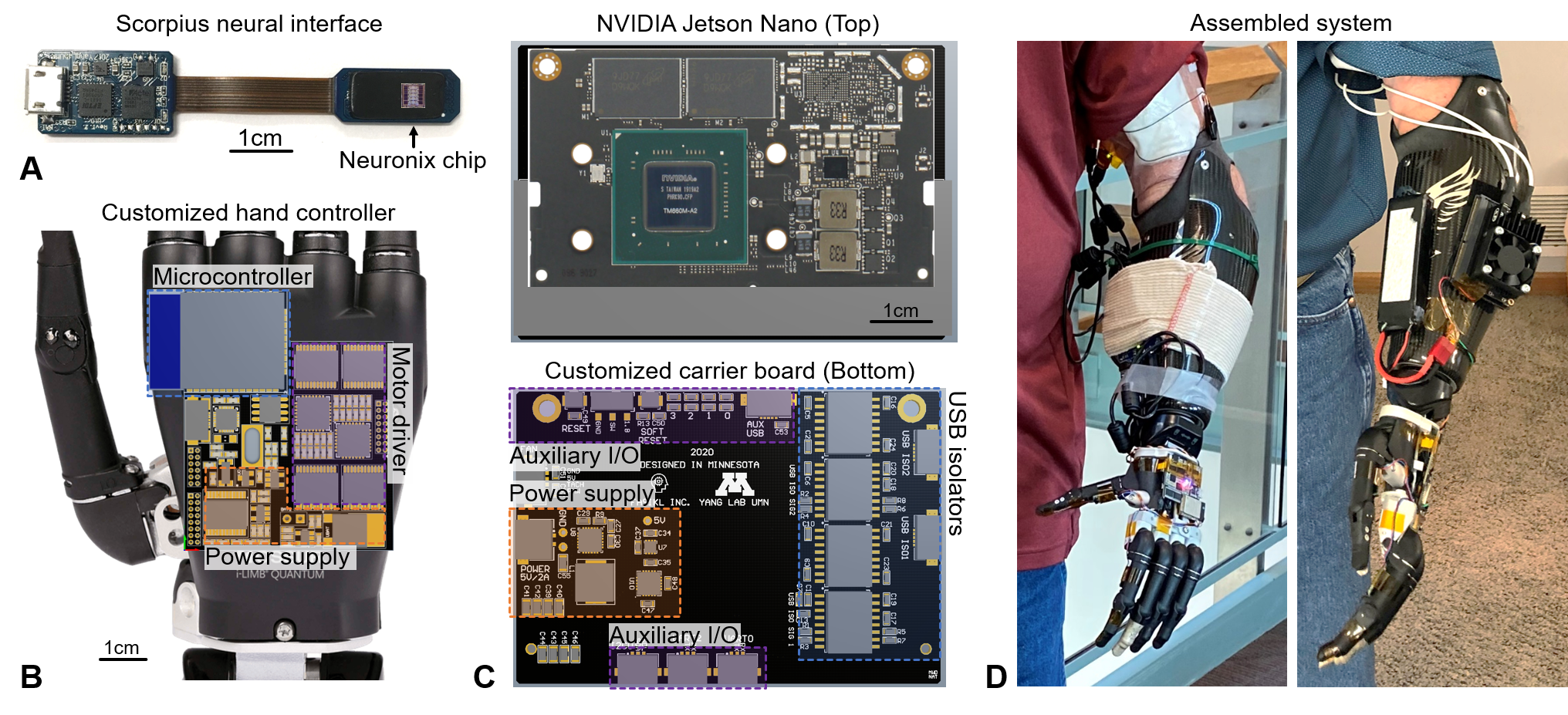}
%\vspace{-10pt}
\caption{Hardware implementation. (A) The Scorpius neural interface. (B) The customized controller within the interior of the i-Limb hand. (C) The NVIDIA Jetson Nano module (heatsink removed) and the customized carrier board. (D) The fully-assembled system is attached to the amputee's existing socket.}
%\vspace{-10pt}
\label{Fig_Hardware}
\end{figure}

Fig. \ref{Fig_Hardware} shows a closer look at the hardware implementation of individual components. Each Scorpius device (Fig. \ref{Fig_Hardware}(A)) is equipped with a 3mm x 3mm Neuronix chip which contains the fully-integrated FS amplifier and ADC. The rest of the device provides low-noise supply voltages for the chip and relays raw nerve data to the AI-engine through a single $\mu$USB connector. The miniaturized form-factor of the Scorpius system allows it to be deployed in various wearable and implantable applications. The device is powered by the carrier board through an isolated USB link.

The Jetson Nano's carrier board (Fig. \ref{Fig_Hardware}(C)) is located directly under the Nano module itself with similar physical dimensions. A 260-pin SODIMM connector provides the interface between the two boards. They are powered by the main Li-ion battery. The power supply circuits provide the main rail (5V, 2A) for the Nano module and other voltages (3.3V, 1.8V). USB isolators based on ADuM5000 and ADuM4160 (Analog Devices, MA, USA) are used to prevent the digital noise from affecting the Scorpius analog front-end performance. Other general-purpose I/O such as UART, I2C, USB (nonisolated), etc., are also provided on the carrier board. 

The customized hand controller board (Fig. \ref{Fig_Hardware}(B)) fits within the interior of the i-Limb hand.  It is powered by the main Li-ion battery. The power supply circuits provide the (3.3V) rail for the ESP32 module and the (12V) rail for the motor driver. The motor driver circuits convert the ESP32 outputs into the PWM signals needed to actuate the DC motors. 

Fig. \ref{Fig_Hardware}(D) presents a look at a fully-assembled system that is attached to the amputee's existing socket. We also show an early prototype that uses the default Jetson Nano's carrier board with off-the-shelf USB isolators and a power bank. While being seemingly bulkier, the early prototype's motor decoding functions and one with customized PCB is identical.

\subsection{Data Processing Flow}
\begin{figure}[ht]
\centering
\includegraphics[trim=10 0 0 0,clip,width=1\linewidth]{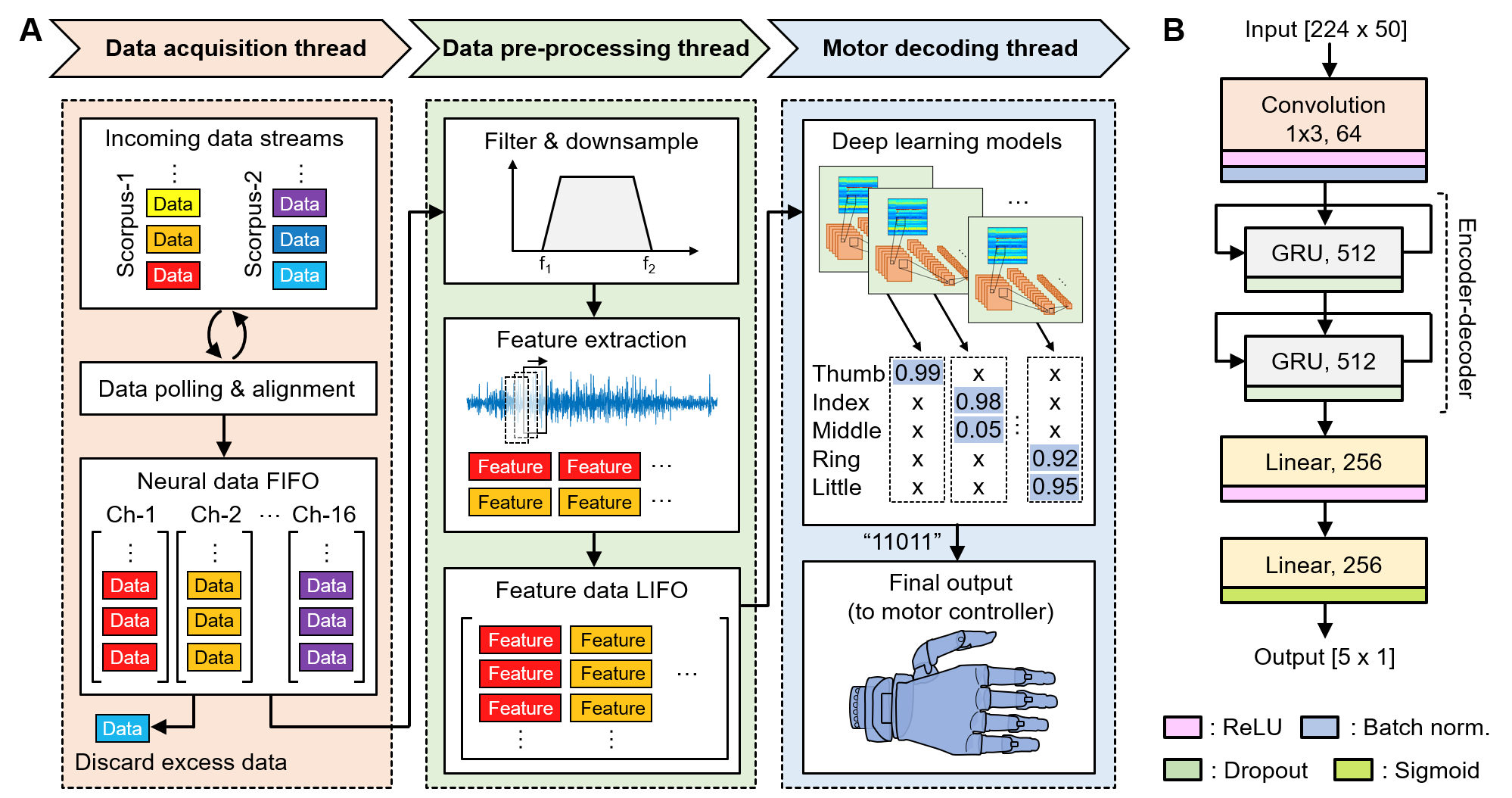}
%\vspace{-5pt}
\caption{Software implementation. (A) Overview of the data processing flow deployed on the Jetson Nano. The program is implemented in Python and consists of three threads. (B) The architecture of the deep learning-based motor decoder. Up to five deep learning models could be deployed, each controls the movement of one or more fingers.}
%\vspace{-10pt}
\label{Fig_Software}
\end{figure}
\begin{table}[ht]
\centering
\caption{List of features for motor decoding}
\vspace{10pt}
\includegraphics[width=1\textwidth]{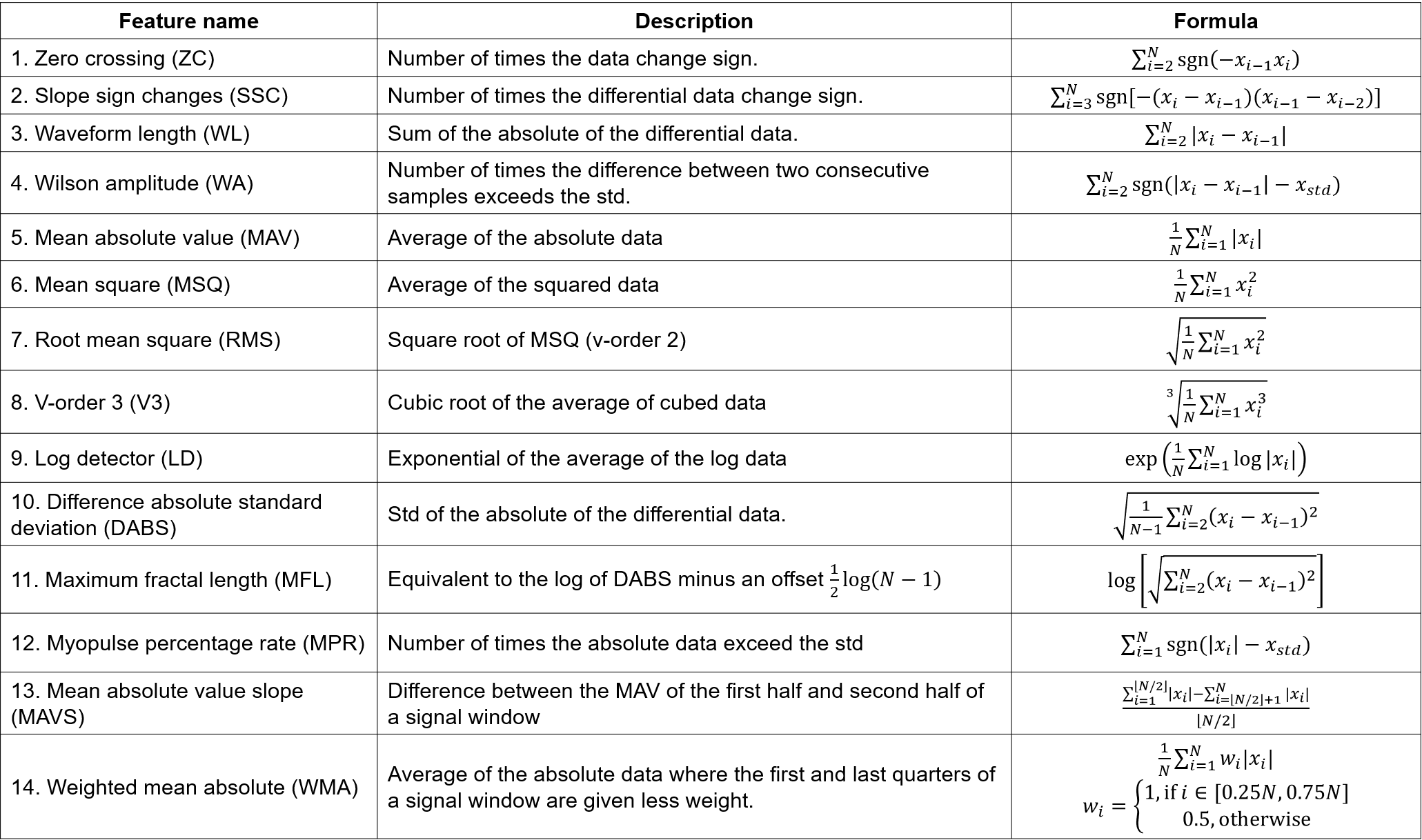}
\vspace{-10pt}
\label{Table_Feature}
\end{table}

Fig. \ref{Fig_Software}(A) shows an overview of the data processing flow deployed on the Jetson Nano. The program is implemented in Python and consists of three separate threads for data acquisition, data pre-processing, and motor decoding. The multi-threading implementation helps maximize the utilization of the quad-core CPU and reduces the processing latency. 

The data acquisition thread polls data from two or more Scorpius devices and aligns them into appropriate channels. The data streams, one for every device, continuously fill up the USB buffers at a bitrate of 1.28Mbps per device. Each stream contains data from eight channels at a sampling rate of 10kHz. Headers are also added to properly separate and align the data bytes into individual channels, which are then placed into first-in-first-out (FIFO) queues for the data pre-processing thread. 

The data pre-processing thread filters, downsamples raw nerve data, and subsequently performs feature extraction according to the procedures outlined in \cite{2021_Luu}. Table \ref{Table_Feature} shows the list, descriptions, and formula of fourteen features used for motor decoding. We utilize nerve data in the 25-600 Hz band, which are shown in \cite{2020_Nguyen} to contain the majority of the signals' power. We apply an anti-aliasing filter at 80\% the Nyquist frequency, downsampling by 2-time, then apply the main 4th-order bandpass filter with 25-600 Hz cut-offs. In the feature extraction task, each feature data point is computed over a sliding window of 100msec with 20msec increments, resulting in an effective feature data rate of 50Hz. The feature data are placed into last-in-first-out LIFO queues (rolling matrices) for the motor decoding thread. Unlike previous studies, no data are stored for offline analysis. Nerve data are processed and fed to the decoder as soon as they are acquired or discarded if the thread cannot keep up. This setup ensures the decoder always receives the latest data. In practice, the buffer to python queue time is negligible, and the pre-processing time is the bulk of the non-motor decoding latency. Excess data could also be caused by a small mismatch in clock frequency between different Scorpius devices, which creates data streams with a higher bitrate than others. As a result, up to 60msec of raw data is occasionally discarded.

The motor decoding thread runs deep learning inference using the most up-to-date feature data from the LIFO queues corresponding to the past 1sec of nerve signals. There are one to five deep learning models; each decodes the movements of one or more fingers. All models have the same architecture, but could be trained on different datasets to optimize the performance of a specific finger. The reason is that while an individual model can produce a [5$\times$1] prediction matrix, it is often difficult to train a single model that is optimized for all five fingers. For example, the first model in Fig. \ref{Fig_Software}(A) only decodes the thumb movement. Because the control signals associating with the thumb are the strongest among the fingers for this particular amputee, the thumb training typically converges in 1-2 epochs. Additional training could cause over-fitting. The final prediction output is sent to the hand controller via a serial link for operating the robotic hand and/or to a remote computer via a Bluetooth connection for debugging.

\subsection{Deep Learning-Based Motor Decoder}

Fig. \ref{Fig_Software}(B) shows the design of the deep learning motor decoder based on the RNN architecture and implemented using the PyTorch\footnote{\url{https://pytorch.org/}} library. The input matrix dimensions are [224$\times$50] = [16 channels] $\times$ [14 features] $\times$ [50 time-step]. Here 50 points of feature data at the effective 50Hz rate correspond to 1sec of past neural data. The output matrix dimensions are [5$\times$1] corresponding to five fingers. The design is modified from \cite{2021_Luu} by adding and removing certain layers while tracking the overall efficacy using the 5-fold cross-validation on a small batch of data.

The initial convolutional layer identifies different representations of data input. The subsequent encoder-decoder utilizes gated recurrent units (GRU) to represent the time-dependent aspect of motor decoding. The two linear layers perform analysis on the decoder's output and produce the final output matrix, which are the probabilities individual finger is active. 50\% dropout layers are added to avoid over-fitting and improve the network's efficiency. Overall, each model consists of 1.6 million parameters in total.
 
The models are trained on a desktop PC with an Intel Core i7-8086K, and an NVIDIA RTX 2080 Super. We use the Adam optimizer \cite{2017_Adam} with the default parameters $\beta_1=0.99$, $\beta_2=0.999$, and a weight decay regularization $L_2 = 10^{-5}$. The mini-batch size is set to 64. The number of epoch (2-10) and initial learning rate ($10^{-4}-10^{-3}$) are adjusted for each model to optimize the performance while preventing over-fitting. The learning rate is reduced by a factor of 10 when the training loss stopped improving for two consecutive epochs. The training time for each epoch depends on the dataset's size and typically takes about 10-15 sec.

%%%%%%%%%%%%%%%%%%%%%%%%%%%%%%%%%%%%%%%%%%%%%%%%%%%%%%%%%%%%%%%%%%%%%%%% 
\section{Motor Decoding Experiment}

\subsection{Human Experiment Protocol}

The human experiment is a part of the clinical trial DExterous Hand Control Through Fascicular Targeting (DEFT), which is sponsored by the DARPA Biological Technologies Office as part of the Hand Proprioception and Touch Interfaces (HAPTIX) program, identifier No. NCT02994160\footnote{\url{https://clinicaltrials.gov/ct2/show/NCT02994160}}.

The human subject is a male, 46-year-old transradial amputee who has lost his hand for 14 years. The human experiment protocols are reviewed and approved by the Institutional Review Board (IRB) at the University of Minnesota (UMN) and the University of Texas Southwestern Medical Center (UTSW). The amputee voluntarily participates in our study and is informed of the methods, aims, benefits, and potential risks of the experiments prior to signing the Informed Consent. Patient safety and data privacy are overseen by the Data and Safety Monitoring Committee (DSMC) at UTSW. The implantation, initial testing, and post-operative care are performed at UTSW by Dr. Cheng and Dr. Keefer, while motor decoding experiments are performed at UMN by Dr. Yang's lab. The clinical team travels with the patient in each experiment session. The patient also completes the Publicity Agreements where he agrees to be publicly identified, including showing his face.

\subsection{Nerve Data Acquisition}
\begin{figure}[p]
\centering
\includegraphics[width=1\textwidth]{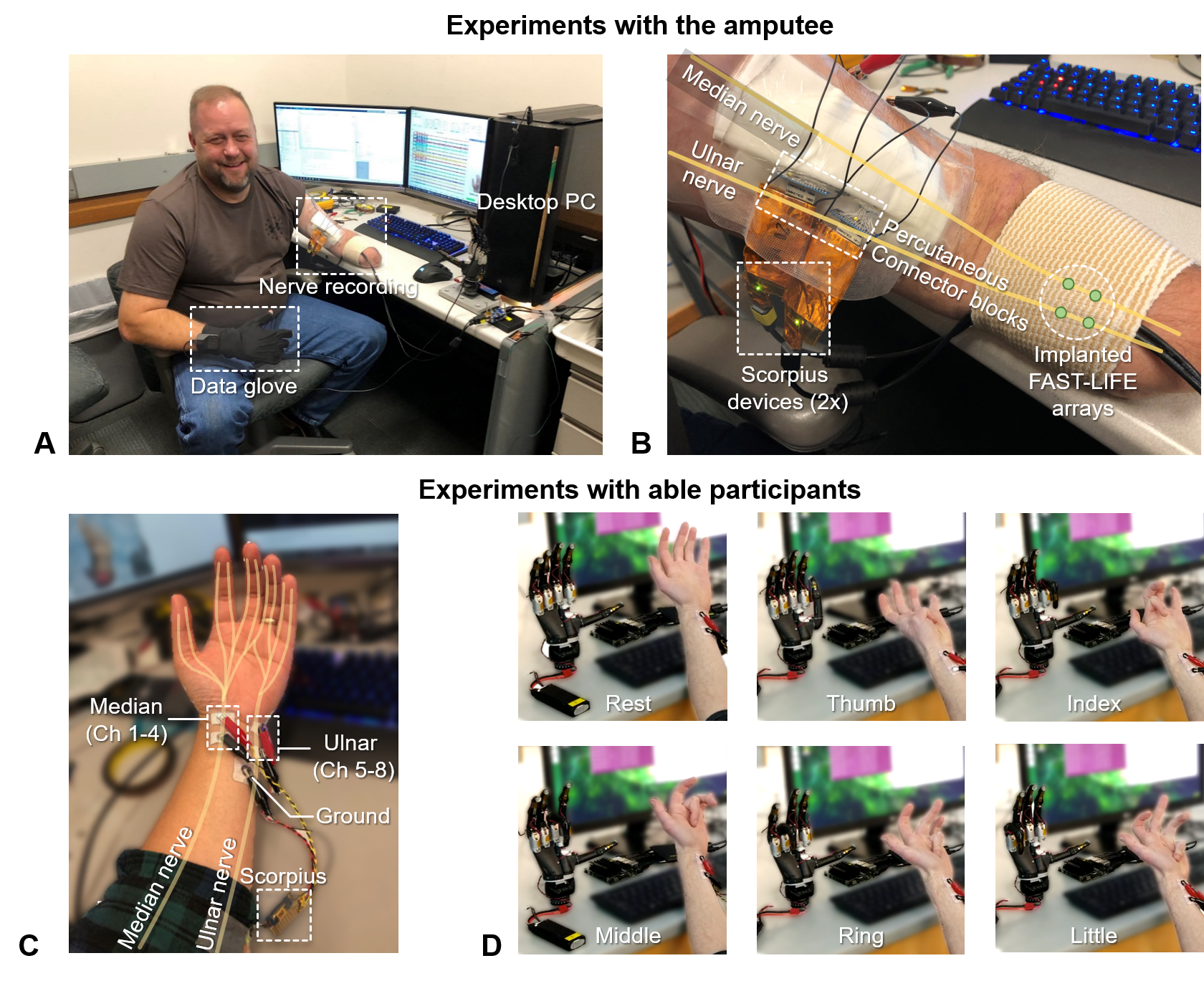}
%\vspace{-10pt}
\caption{Motor decoding experiment setup. (A) The amputee during a training session using the mirror movement paradigm. (B) Zoom-in of the residual limb showing the approximate microelectrode implants and nerve data acquisition using two Scorpius devices. (C, D) Additional experiments with able participants to verify the system's functions where the prothetic hand mimics the participant's finger movements.}
%\vspace{-10pt}
\label{Fig_ExpSetup}
\end{figure}

The main experiment with the amputee uses nerve signals acquired with intrafascicular microelectrodes as shown in Fig. \ref{Fig_ExpSetup}(A, B). The patient undergoes an implant surgery where four longitudinal intrafascicular electrode (LIFE) arrays are inserted into the residual median and ulnar nerves using the microsurgical fascicular targeting (FAST) technique. Within one nerve, the arrays are placed into two discrete fascicle bundles. More details regarding the microelectrode arrays' design and characteristics and the surgical procedures are reported in \cite{2017_Cheng, 2019_Overstreet}. 

Fig. \ref{Fig_ExpSetup}(A) shows the amputee with the experiment setup for acquiring training data. Fig. \ref{Fig_ExpSetup}(B) shows a zoom-in of the patient's injured hand. We mark the implants' approximate location near the end of the stump, which strategically locates far away from residual muscles in the forearm to minimize volume-conducted EMG. The electrode wiring is brought out through percutaneous holes on the arm and secured to two connector blocks. The Scorpius devices attach to the blocks via two standard 40-pin Omnetics nano connectors. Extensive analysis of the nerve recordings is provided in \cite{2020_Nguyen}. The results suggest ENG signals associated with voluntary compound action potentials (vCAP) and single-axon action potentials (spikes) are the acquired nerve data's primary components. We also show a strong correlation between nerve activities and finger movements and clear differentiation between the median and ulnar nerves, which is consistent with human anatomy.

In addition, we perform motor decoding with able participants using surface recordings as shown in Fig. \ref{Fig_ExpSetup}(C, D). The experiment's purpose is to verify the system's key functions and evaluate the overall performance by providing the researchers an actual ``feeling'' of the prosthesis's movements. It is an indispensable testbed for us to replicate, debug and optimize the motor decoding paradigm without the amputee, which is essential to avoid wasting the patient's valuable experimenting time.

Fig. \ref{Fig_ExpSetup}(A) shows the experiment setup with able participants using wet-gel Ag/AgCl electrodes (1cm x 2cm) to obtain data from the median and ulnar nerve. The electrodes are placed along the wrist area, where two nerves run nearest to the skin's surface. The bipolar electrodes are spaced approximately 1cm apart. The acquired data are mixtures of both ENG signals from the two nerves and EMG signals from nearby hand and forearm muscles. Because able participants have strong ENG/EMG signals with all intact muscles, merely two channels are needed to decode the finger movements. We combine channels 1-4 and 5-8 on one Scorpius device to acquire data from the median and ulnar electrodes, respectively. Another device is included to match the 16-channel input from the amputee setup but is kept unattached and only records random noises. Fig. \ref{Fig_ExpSetup}(B) shows a test where the prothetic hand mimics the participant's finger movements.

\subsection{Mirrored Bilateral Training}

The training dataset is acquired with a desktop PC using the mirrored bilateral paradigm similar to \cite{2020_Nguyen}. Nerve signals are obtained from the injured hand with the Scorpius system, while labeled ground-truth movements are captured from the able hand using a data glove. In each experiment session, the participant is instructed to flex a hand gesture 10 times, where the fingers are held in the flexing position in about 2sec. The data glove measures the angle of the finger's proximal phalanx with respect to its metacarpal bone. The data are then thresholded to produce the ground-truth labels for classification. For able participants, the training only includes the flexing of individual fingers to verify the decoder's functions. For the amputee, we add different gestures where two or more fingers are engaged, such as fist/grip (11111), index pinch (11000), pointing (10111), and Hook 'em Horns (10110).

We collect at least four or more mirrored bilateral sessions for each hand gesture. Within a session, the patient does the gesture at different shoulder, arm, and body postures, resembling real-life conditions. Additional sessions could be required for gestures that are difficult to predict. The last data session, which contains the most up-to-date nerve data, is always used for validation while the remaining are used for training. This configuration translates to a training-to-validation ratio of approximately 75:25 to 85:15 for able participants and the amputee, respectively.

%%%%%%%%%%%%%%%%%%%%%%%%%%%%%%%%%%%%%%%%%%%%%%%%%%%%%%%%%%%%%%%%%%%%%%%% 
\section{Experiment Results}

\subsection{Data Processing Latency}
\begin{figure}[t]
\centering
\includegraphics[width=0.9\textwidth]{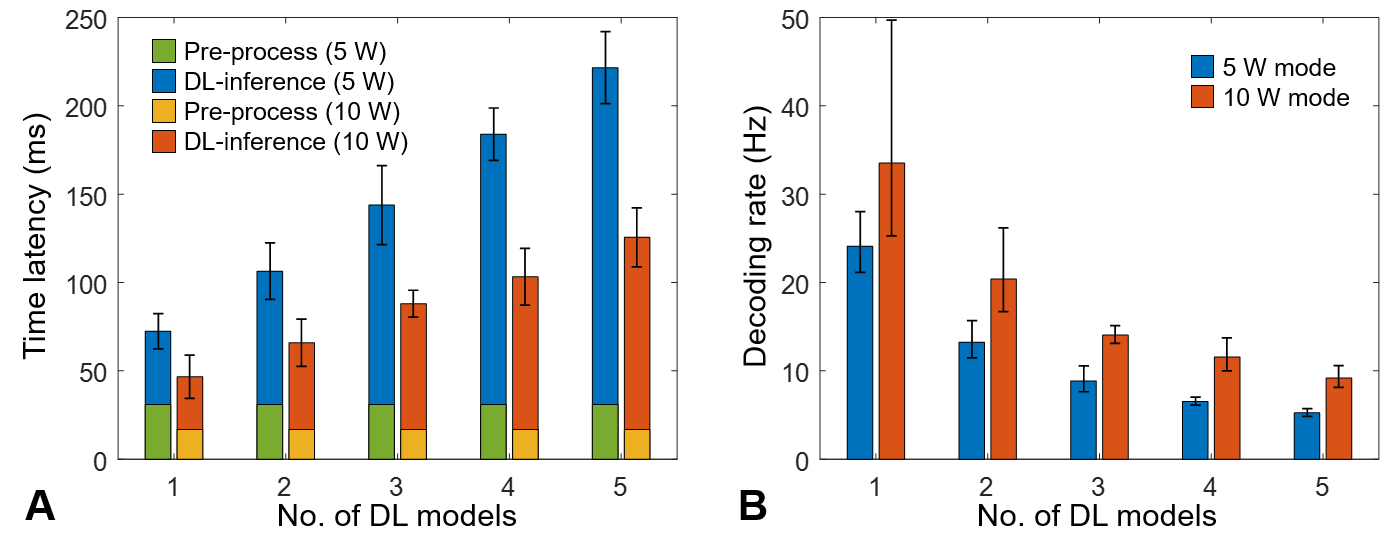}
%\vspace{-10pt}
\caption{Real-time performance metrics. (A) Overall time latency calculated from the moment nerve data are acquired until the corresponding prediction is produced (i.e., input lag). (B) Maximum throughput of the motor decoding pipeline (i.e., frame rate).}
%\vspace{-10pt}
\label{Fig_DecRate}
\end{figure}

Fig. \ref{Fig_DecRate}(A) shows the overall time latency calculated from data acquisition to final classification prediction (i.e., input lag). This metric determines the system's responsiveness and is essential for real-time operation. The latency mainly consists of the data pre-processing and the deep learning-inference steps. Only the latency of deep learning-inference increases linearly as more deep learning models are used. The overall processing time is imposed by the CPU and GPU's clock speed which is subsequently limited by the power budget. Switching the Jetson Nano's power mode from 5W to 10W cuts down the latency approximately by half.

Fig. \ref{Fig_DecRate}(B) shows the maximum throughput of the motor decoding pipeline, which measures the number of predictions produced per second (i.e., frame rate). This metric also contributes to the system's responsiveness. We achieve a decoding rate significantly higher than the inverse of the time latency thanks to the multi-threading implementation. Like the latency, the decoding rate is also decided by the number of deep learning models and the Jetson Nano's power budget.

\subsection{Classification Performance}

\begin{figure}[p]
\centering
\includegraphics[width=1\textwidth]{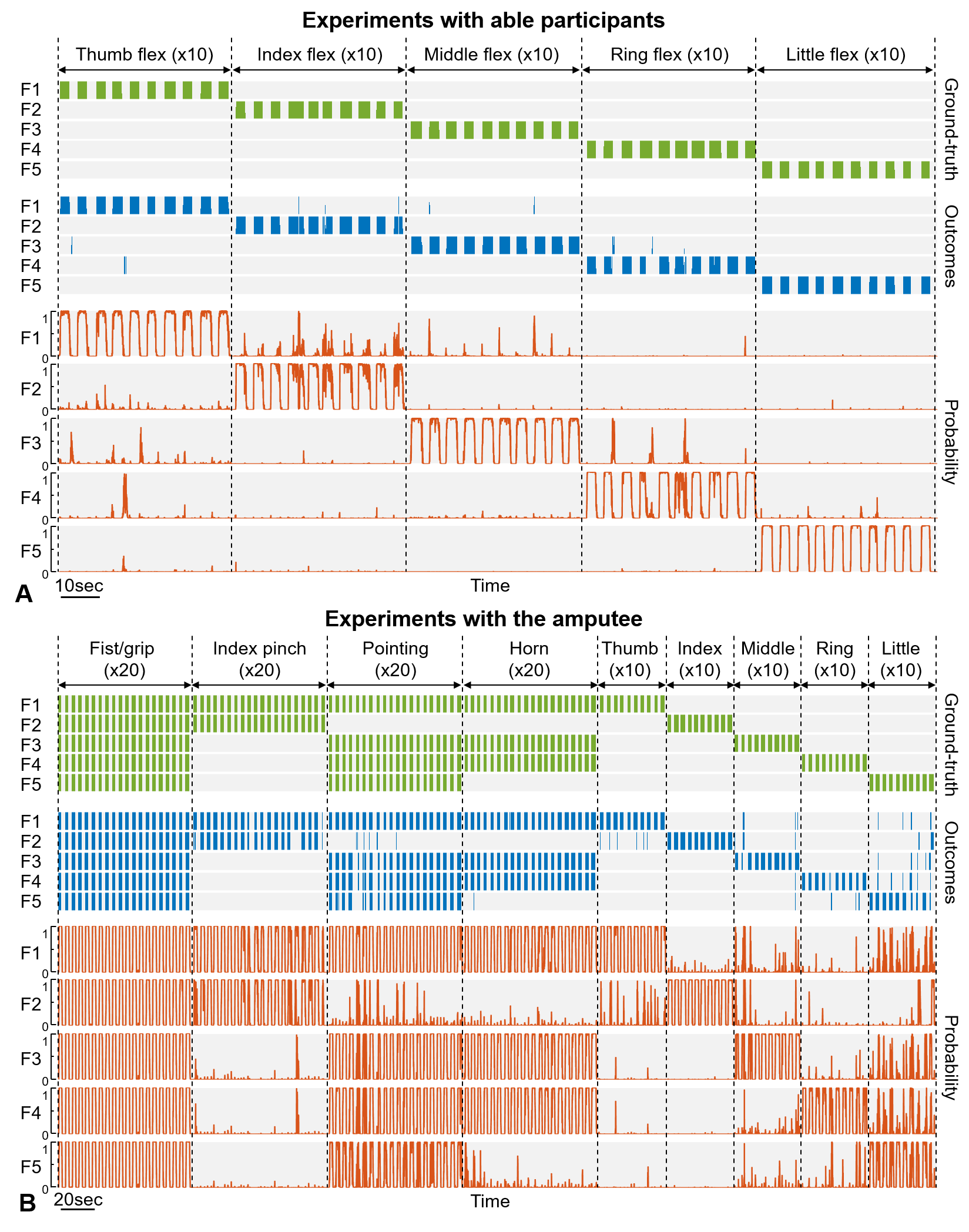}
\vspace{-10pt}
\caption{Classification results from motor decoding experiments with (A) able participants and (B) the amputee show high accuracy prediction could be achieved.}
%\vspace{-10pt}
\label{Fig_Classification}
\end{figure}

Fig. \ref{Fig_Classification}(A, B) shows the classification results, including the prediction outcomes and probability computed over the validation datasets. The size of the training/validation sets are approximately 30,000:10,000 and 150,000:26,000 for able participants and the amputee, respectively. We use one model to predict all five fingers. The prediction results demonstrate that deep learning neural decoders could accurately predict individual fingers' movement, forming distinct hand gestures in the time series.

The classification task's quantitative performance is evaluated using standard metrics, including the true positive rate (TPR) or sensitivity, true negative rate (TNR) or specificity, accuracy, and area under the curve (AUC) derived from true-positive (TP), true-negative (TN), false-positive (FP), and false-negative (FN) as follows:
\begin{gather}
	\text{TPR (sensitivity)} = \text{TP}/(\text{TP}+\text{FN}) \\
	\text{TNR (specificity)} = \text{TN}/(\text{TN}+\text{FP}) \\
	\text{Accuracy} = (\text{TP}+\text{TN})/(\text{TP}+\text{TN}+\text{FP}+\text{FN})
\end{gather}

Table \ref{Table_FuncTest} and Table \ref{Table_HumanExp} show the classification performance results for individual fingers from able participants and the amputee respectively. The data further demonstrate the neural decoder's exceptional capability with accuracy ranging from 95-99\% and AUC ranging from 97-99\%. Nevertheless, there are considerably more false-positives in the amputee prediction compared to able participants. Some finger such as the amputee's index is considerably less predictive than others. This disparity is often caused by a low signal-to-noise ratio in the nerve recordings. In practice, we could either add additional training sessions or train a specific model to control only this particular finger.
\begin{table}[ht]
\centering
\caption{Classification performance from able participants}
\vspace{10pt}
\includegraphics[width=0.6\textwidth]{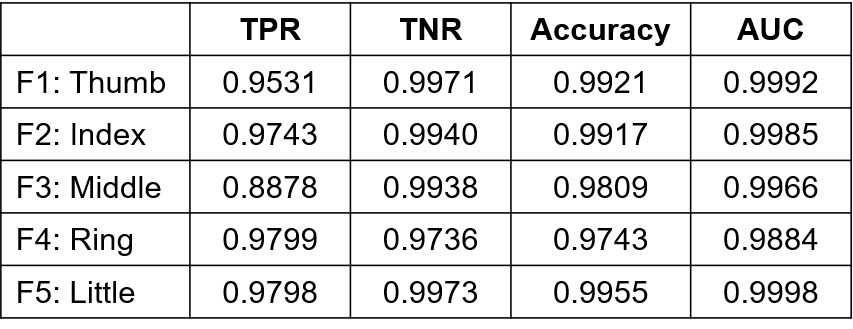}
\vspace{-10pt}
\label{Table_FuncTest}
\end{table}
\begin{table}[ht]
\centering
\caption{Classification performance from the amputee}
\vspace{10pt}
\includegraphics[width=0.6\textwidth]{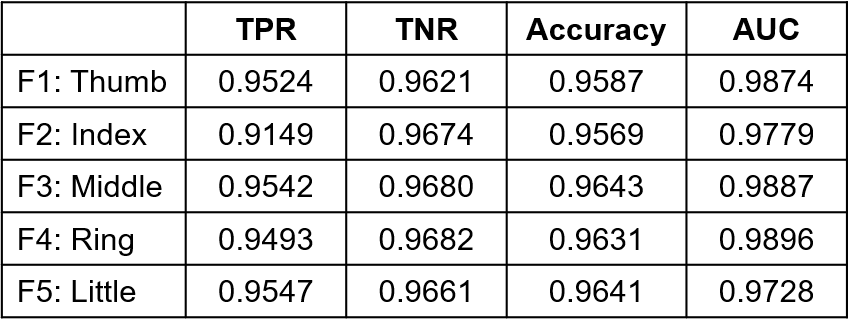}
\vspace{-10pt}
\label{Table_HumanExp}
\end{table}

\subsection{Lab Experiments and Field Tests with The Amputee}
\begin{figure}[p]
\centering
\includegraphics[width=0.95\textwidth]{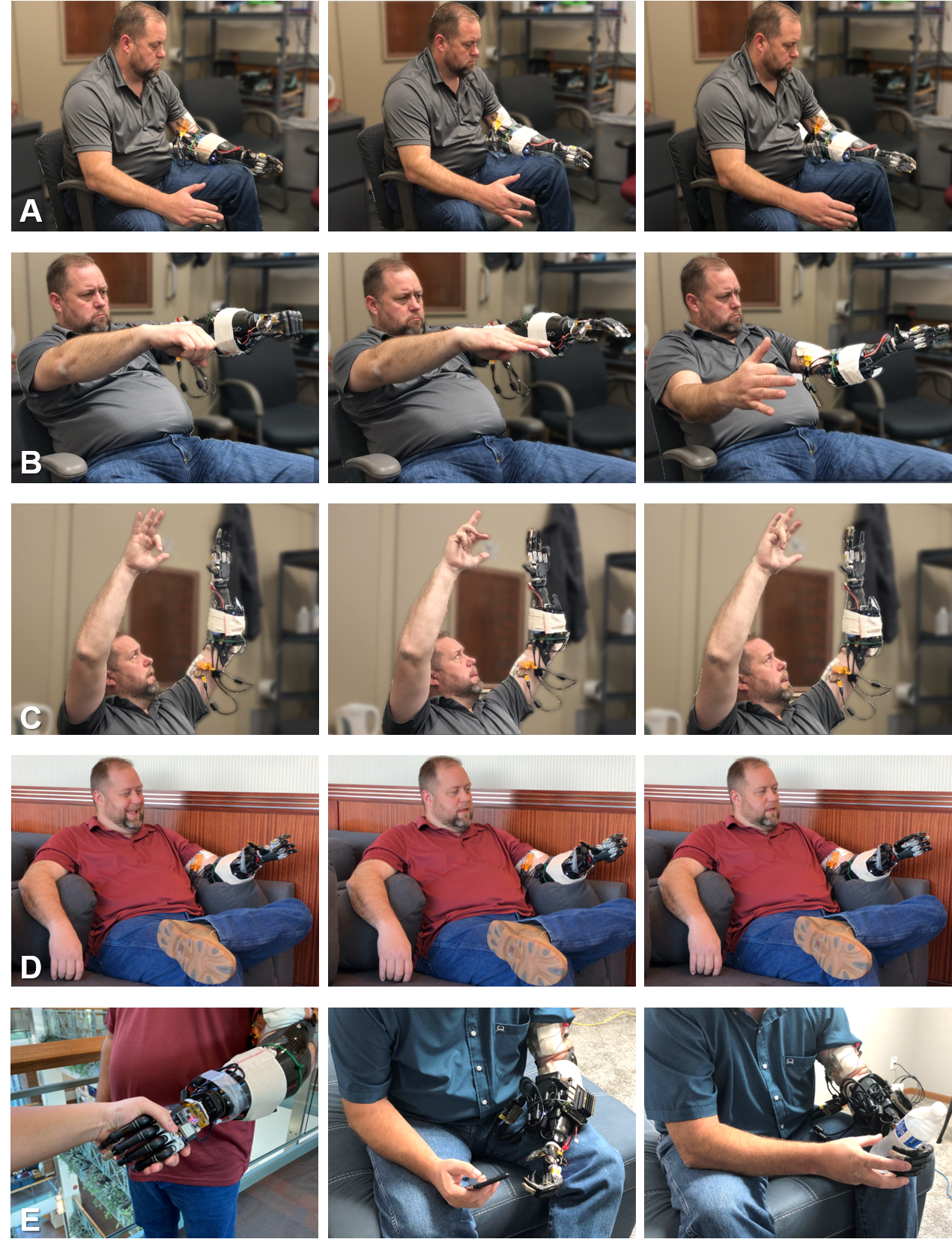}
%\vspace{-10pt}
\caption{(A, B, C) The patient verifies the individual finger control at various arm positions in the lab setting. Here the patient uses his able hand to show outside observers his true motor intention. (D, E) The prosthetic hand is tested in various real-world settings. The system functions remain relatively accurate and robust regardless of the arm/body postures and environments.}
%\vspace{-10pt}
\label{Fig_FieldTest}
\end{figure}

The trained models are loaded onto the Jetson Nano for inference-only operation. The prediction outputs are directly mapped to the movements of the prosthesis's digits. It is worth noting that the motorized fingers move at a much slower rate than the human's hand, thus cannot truly follow the operator's movements. This constraint is a mechanical limitation of the existing prosthesis. Such constraint does not apply when controlling a virtual hand (e.g., MuJoCo) like the experiment in \cite{2020_Nguyen}.

Fig. \ref{Fig_FieldTest}(A, B, C) shows the amputee testing the prosthesis in the laboratory setting. The neural decoder models are trained and optimized on the dataset that is collected about two months earlier. All the data acquisition and processing are made in real-time by the Jetson Nano. There is no wired or wireless communication with any remote computer. The amputee uses his able hand to solely show outside observers his intention of moving the fingers for comparison. The results demonstrate that the robotic hand accurately resembles the operator's motor intent. The amputee also tests the prosthetic hand's robustness at various postures as holding the arm straight out and up could introduce considerable EMG noises. The subject reports a slight change in the system's responsiveness, but there is no significant motor decoding accuracy degradation.

Fig. \ref{Fig_FieldTest}(D, E) shows the patient testing the prosthesis at the building's lobby and lounge. There are various additional noise sources in these real-world settings that could affect the system's functions, such as WiFi, cellphone, electrical appliances, etc. Nevertheless, we do not find any evidence suggesting any significant impact on the system's performance during several hours of continuous operation. Here is the amputee's testimonial during the experiment:
\begin{quote}
[Patient]: I feel like once this thing is fine-tuned as finished products that are out there. It will be more life-like functions to be able to do everyday tasks without thinking of	what positions the hand is in or what mode I have the hand programmed in. It's just like if I want to each and pick up something, I just reach and pick up something. [...] Knowing that it's just like my [able] hand [for] everyday functions. I think we'll get there. I really do!
\end{quote}

\section{Discussion \& Future Work}

\subsection{Software \& Hardware Optimization}

There is still room to improve upon the system implementation. For example, a multithreaded C-like implementation would avoid both the Python global interpreter lock and the runtime interpretation of Python code. The feature extraction implementation also requires further optimization so that overlapping data windows are not computed twice. Another direction is to translate the entire pre-processing thread, including filtering and feature extraction, to a field-programmable gate array (FPGA), significantly reducing the time latency and power consumption. 

The neural network could further be optimized through the use of bit-quantization. For example, the NVIDIA TensorRT is built to optimize network inference on CUDA devices, and the major libraries have quantization tools for 8-bit integer or 16-bit floating-point inference. However, these options have limitations in network-layer support, as quantization of recurrent network layers is still an active research topic. In the current implementation, 8-bit integer quantization could not be done due to incompatibilities in CUDA-quantization support for PyTorch and model-to-model conversion restrictions regarding recurrent network layers. These optimizations will likely become necessary to handle networks trained for higher degrees of freedom, due to the likelihood that multiple networks will need to be used and/or these networks will contain higher parameter counts.

The Jetson Nano's GPU is based on NVIDIA's Maxwell (2014) architecture which may not be as efficient as more recent GPU architectures like Turing (2018) and Ampere (2020). It would also be possible to combine the Jetson Nano with USB-attached devices like the Intel Compute Stick and Google Coral to extend the deep learning inference capability once the software support becomes more robust.

\subsection{Online Motor Decoding Optimization}

Unlike our previous works \cite{2020_Nguyen, 2021_Luu}, here we do not perform regression of the finger movements due to the constraint of the existing portable platform. This setup limits the dexterity and the range of the prosthesis movements, e.g., stopping the finger halfway. To support regression, the deep learning models and/or the system implementation need to be further optimized to retain real-time operation.

The predictability of individual hand gestures largely depends on the specific patient and the microelectrode implant. Fig. \ref{Fig_Classification}(B) only includes gestures yielding the best results when training the deep learning models. The correlation among different gestures is also relatively complex. For example, the training process indicates that the ``fist/grip'' gesture is the most predictable, which converges in only 1-2 epoch; however, the model then struggles to differentiate the ``fist/grip'' from the ``thumb up'' gesture. Feasible approaches include fine-tuning the model's architecture and adding more training data with various holding patterns for each hand gesture.

The amount of past nerve data needed to decode the movement may need further investigation. Longer data significantly increase the size of the deep learning models but could yield better prediction accuracy. The offline analysis in \cite{2020_Nguyen, 2021_Luu} uses 4sec of past data, while the online implementation here only uses 1sec of past data. In general, both the able participants and the amputee report that shorter data makes the system more responsive; however, too short (e.g., 0.2sec) and finger movements become ``jerky''.

Furthermore, there in an open question regarding what is considered to be an appropriate latency. Our current system achieves an ``acceptable'' latency of 50-120 msec (Fig. \ref{Fig_DecRate}(A)), which is computed from the moment nerve data are acquired to the moment the corresponding motorized finger starts to move. Here the neural decoder emulates a processor for determining the kinematics required to perform a motor action; thus the ideal decoding latency can be approximated to motor intent delay in human motor control circuits. While an exact response time can be up to 200msec dependent on the type of task, stimulus input, and definition of motor delay \cite{2015_Wong}, a central-processing delay of 16 to 26 msec according to Fitts’ (1954) reciprocal finger-tapping task has been referred to as a ``typical'' motor intent delay \cite{2008_Beamish}.

\section{Conclusion}

This paper investigates the feasibility of deploying deep learning for neuroprosthesis applications using edge computing platforms like the NVIDIA Jetson Nano. We implement a portable, self-contained system that could be attached to the existing prosthetic socket. It allows an amputee to control individual finger movements in real-time using peripheral nerve recordings. Testings show the system is robust in various arm/body postures, as well as in different laboratory and real-world environments during several hours of continuous operation. We envision this work as a stepping stone to future deployments of artificial intelligence for portable biomedical devices.

\section*{Acknowledgments}

The surgery and patients related costs were supported in part by the DARPA under Grants HR0011-17-2-0060 and N66001-15-C-4016. The human motor decoding experiments, including the development of the prototype, was supported in part by MnDRIVE Program and Institute for Engineering in Medicine at the University of Minnesota, in part by the NIH under Grant R01-MH111413-01, in part by NSF CAREER Award No. 1845709, in part by Fasikl Incorporated, and in part by Singapore Ministry of Education funding R-263-000-A47-112, and Young Investigator Award R-263-000-A29-133.

Zhi Yang is co-founder of, and holds equity in, Fasikl Incorporated, a sponsor of this project. This interest has been reviewed and managed by the University of Minnesota in accordance with its Conflict of Interest policy. J. Cheng and E. W. Keefer have ownership in Nerves Incorporated, a sponsor of this project.

\end{document}